\documentclass{article}

\usepackage{times}
\usepackage{epsfig}
\usepackage{graphicx}
\usepackage{amsmath}
\usepackage{amssymb}
\usepackage{bm}
\usepackage{multirow}
\usepackage{arxiv}
\usepackage[utf8]{inputenc} 
\usepackage[T1]{fontenc}    
\usepackage{url}            
\usepackage{booktabs}       
\usepackage{amsfonts}       
\usepackage{nicefrac}       
\usepackage{microtype}      
\usepackage{caption}
\usepackage{subfig}
\usepackage{tikz}
\usepackage{pgfplots}

\graphicspath{ {images/} }

\def\ie{\emph{i.e.~}}
\def\eg{\emph{e.g.~}}
\def\etc{\emph{etc}}

\title{Scale-Aware Multi-Level Guidance for Interactive Instance Segmentation}

\author{
  Soumajit Majumder \\
  Institute of Computer Science II\\
  University of Bonn\\
  \texttt{majumder@cs.uni-bonn.de} \\
   \And
  Angela Yao \\
  School of Computing\\
  National University of Singapore\\
  \texttt{yaoa@comp.nus.edu.sg} \\
}

\begin{document}
\maketitle

\begin{abstract}
In interactive instance segmentation, users give feedback to iteratively refine segmentation masks. The user-provided clicks are transformed into guidance maps which provide the network with necessary cues on the whereabouts of the object of interest. Guidance maps used in current systems are purely distance-based and are either too localized or non-informative. We propose a novel transformation of user clicks to generate scale-aware guidance maps that leverage the hierarchical structural information present in an image. Using our guidance maps, even the most basic FCNs are able to outperform existing approaches that require state-of-the-art segmentation networks pre-trained on large scale segmentation datasets. We demonstrate the effectiveness of our proposed transformation strategy through comprehensive experimentation in which we significantly raise state-of-the-art on four standard interactive segmentation benchmarks.
\end{abstract}

\section{Introduction}
Interactive object selection and segmentation allows users to interactively select objects of interest down to the pixel level by providing inputs such as clicks, scribbles, or bounding boxes. The segmented results are useful for downstream applications such as image/video editing~\cite{vos-wild,lazysnap}, image-based medical diagnosis~\cite{med2,deepigeos}, human-machine collaborative annotation~\cite{fluid}, and so on. GrabCut~\cite{grabcut} is a pioneering example of interactive segmentation which segments objects from a user-provided bounding box by iteratively updating a colour-based Gaussian mixture model. Other methods include Graph Cuts~\cite{graphcuts}, Random Walk~\cite{randomwalk} and GeoS~\cite{geos} though more recent methods~\cite{risnet,itis,dextr,deepgc,ifcn} approach the problem with deep learning architectures such as convolutional neural networks (CNNs).

\begin{figure}[t!]
	\begin{center}
	\includegraphics[trim=0cm 14.0cm 0cm 0cm,clip,width=0.8\linewidth]{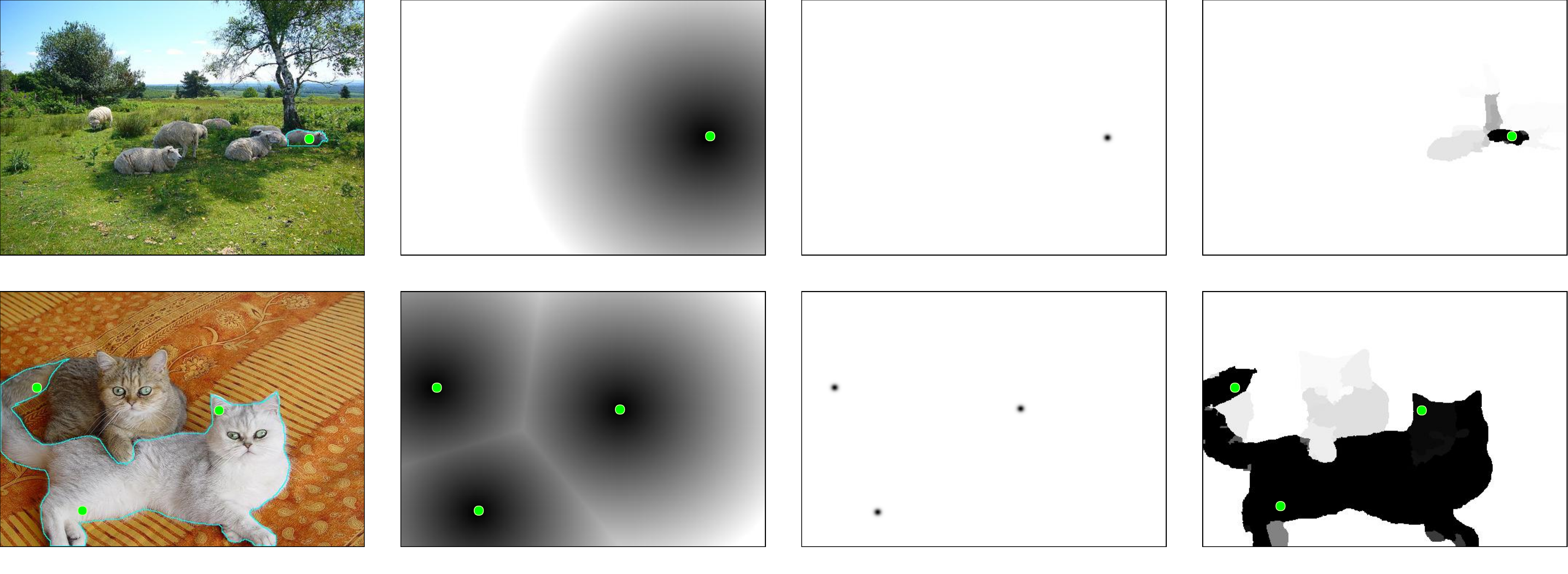}
	\end{center}
	\caption{\textbf{Motivation.} Existing interactive instance segmentation~\cite{ifcn} techniques do not utilize any image information when generating \emph{guidance} maps (second and third column). In contrast, our proposed technique exploits image structures such as super-pixels and object proposals, allowing us to generate much more informative guidance maps (final column).} 
	\label{fig:motivation}
\end{figure}

In standard, non-interactive instance segmentation~\cite{watershed,instanceaware,sds,hypercolumns,maskrcnn}, the RGB image is given as input and segmentation masks for each object instance are predicted. In an interactive setting, however, the input consists of the RGB image as well as \emph{`guidance'} maps based on user-provided supervision.  The guidance map helps to select the specific instance to segment; when working in an iterative setting, it can also help correct errors from previous  segmentations~\cite{vos-wild,itis,risnet,ifcn}.

User feedback is typically given in the form of either clicks~\cite{risnet,itis,dextr,ifcn,twostream,latentdiversity} or bounding boxes~\cite{deepgc} and are then transformed into guidance signals fed as inputs into the CNN. By working with high-level representations encoded in pre-trained CNNs, the number of user interactions required to generate quality segments have been greatly reduced.  However, there is still a large incongruence between the image encoding versus the guidance signal, as user interactions are transformed into simplistic primitives such as Euclidean~\cite{ifcn,latentdiversity,twostream} or Gaussian distance maps~\cite{vos-wild,itis,dextr} (Fig.~\ref{fig:motivation} second and third row respectively), the latter being the preferred choice in more recent works due to their ability to localize user clicks~\cite{itis}.

Our observation is that current guidance signals disregard even the most basic image consistencies present in the scene, such as colour, local contours, and textures.  This of course also precludes even more sophisticated structures such as object hypotheses, all of which can be determined in an unsupervised way.  As such, we are motivated to maximize the information which can be harnessed from user-provided clicks and generate more meaningful guidance maps for interactive instance segmentation.

To that end, we propose a simple yet effective transformation of user clicks which enables us to leverage a hierarchy of image information, starting from low-level cues such as appearance and texture, based on superpixels, to more high-level information such as class-independent object hypotheses (see Fig.~\ref{fig:samplemaps}).  

Ours is the first work to investigate the impact of guidance map generation for interactive segmentation.  Our findings suggest that current Gaussian- and Euclidean distance based maps are too simple and do not fully leverage structures present in the image. A second and common drawback of current distance-based guidance maps is that they fail to account for the scale of the object during interaction. Object scale has a direct impact on the network performance when it comes to classification~\cite{explicitscale} or segmentation~\cite{learndeconv}. Gaussian- and Euclidean distance maps are primarily used for localizing the user clicks and do not account for the object scale. Our algorithm roughly estimate object scale based on the user-provided clicks and refines the guidance maps accordingly. 

Our approach is extremely flexible in that the generated guidance map can be paired with any method which accepts guidance as a new input channel~\cite{ifcn,risnet,itis,vos-wild}. We demonstrate via experimentation that providing scale-aware guidance by leveraging the structured information in an image leads to a significant improvement in performance when compared to the existing state-of-the-art, all the while using a simple, off-the-shelf, CNN architecture. The key contributions of our work are as follows : 

\begin{itemize}
	\item We propose a novel transformation of user-provided clicks which generates guidance maps by leveraging hierarchical information present in a scene.  
	\item We propose a framework which can account for the scale of an object and generate the guidance map accordingly in a click-based user feedback scheme.
	\item We perform a systematic study of the impact of guidance maps on the interactive segmentation performance when generated based on features at different image hierarchy. 		
	\item We achieve state-of-the-art performance on the Grabcut, Berkeley, Pascal VOC 2012 and MS COCO datasets. Our proposed method significantly reduces the amount of user interaction required for accurate segmentation and uses the fewest number of average clicks per instance.
\end{itemize}

\section{Related Works}

Segmenting objects interactively using clicks, scribbles, or bounding boxes has always been a problem of interest in computer vision research, as it can solve some quality problems faced by fully-automated segmentation methods.  Early variants of interactive image segmentation methods, such as the parametric active contour model~\cite{snakes} and intelligent scissors~\cite{scissors} mainly considered boundary properties when performing segmentation; as a result they tend to fare poorly on weak edges. More recent methods are based on graph cuts~\cite{graphcuts,grabcut,growcut,lazysnap}, geodesics~\cite{geodesicmatting,geos}, and or a combination of the two~\cite{geodesic,geodesicgraphcut}. However, all these algorithms try to estimate the foreground/background distributions from low-level features such as color and texture, which are unfortunately insufficient in several instances, \eg  in images with similar foreground and background appearances, intricate textures, and poor illumination.

As with many other areas of computer vision, deep learning-based methods have become popular also in interactive segmentation in the past few years. In the initial work of~\cite{ifcn}, user-provided clicks are converted to Euclidean distance transform maps which are concatenated with the color channels and fed as input to a FCN~\cite{fcn}. Clicks are then added iteratively based on the errors of the previous prediction. On arrival of each new click, the Euclidean distance transform maps are updated and inference is performed. The process is repeated until satisfactory results is obtained. Subsequent works have focused primarily on making extensions with newer CNN architectures~\cite{itis,vos-wild} and iterative training procedures~\cite{itis,risnet}.  In the majority of these works, user guidance has been provided in the form of point clicks~\cite{ifcn,itis,risnet,dextr,latentdiversity} which are then transformed into a Euclidean-based distance map~\cite{ifcn,latentdiversity}.  One observation made in~\cite{vos-wild,itis,dextr} was that encoding the clicks as Gaussians led to some performance improvement because it localizes the clicks better~\cite{itis} and can encode both positive and negative click in a single channel~\cite{vos-wild}. In~\cite{tapnshoot}, the authors explore the use of superpixels to generate the guidance map. In contrast to~\cite{tapnshoot}, our guidance map is able to incorporate image features at different scales and without requiring an elaborate graph-based optimization for the generation of the guidance map. For the most part, there has been little attention paid to how user inputs should be incorporated as guidance; the main focus in interactive segmentation has been dedicated towards the training procedure and network architectures. 

\section{Proposed Approach}

\begin{figure*}[t]
	\begin{center}
	\includegraphics[width=0.95\linewidth]{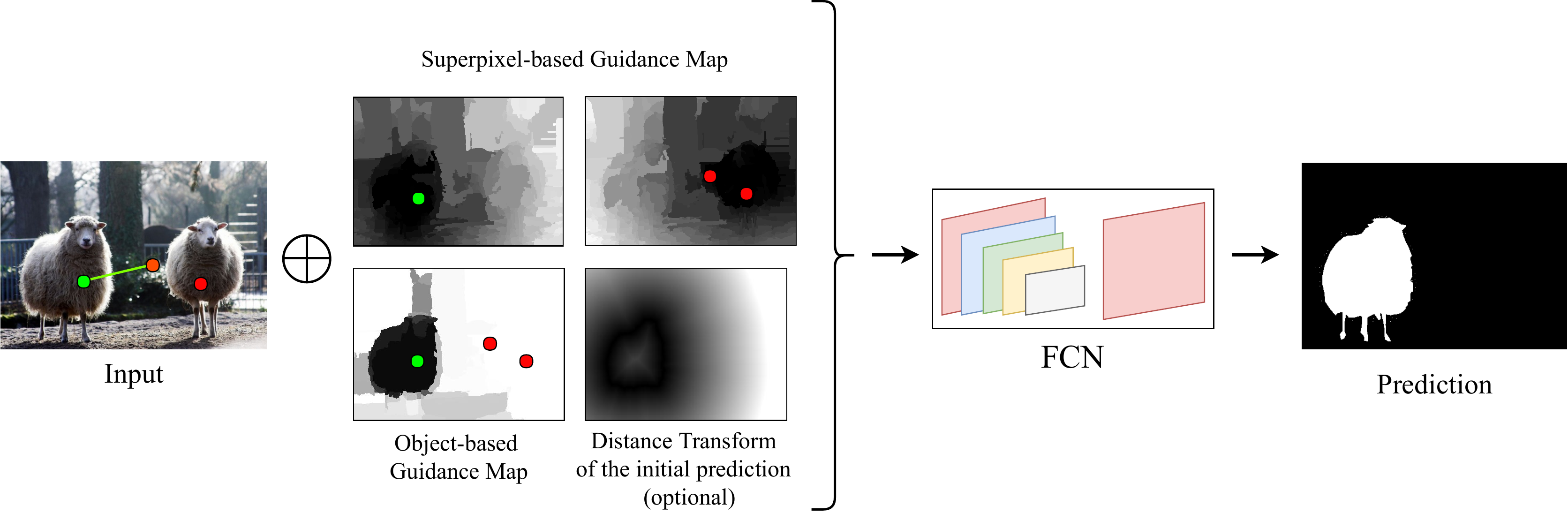}
	\end{center}
	\caption{\textbf{Outline of our method.} Given an input image and user interactions, we transform the positive and negative clicks (denoted by the green and red dots respectively) into three separate channels ($2$ channel superpixel-based and $1$ object proposal-based guidance map), which are concatenated (denoted as $\oplus$) with the $3$-channel image input and is fed to our network. Additionally, we concatenate the euclidean distance transform of the predicted mask from the previous iteration as our final non-color channel. The solid green line indicates our estimate of the object scale based on the initial pair of positive and negative click. The output is the ground truth map of the selected object. }
	\label{fig:outline}
\end{figure*}

We follow previous interactive frameworks~\cite{ifcn,risnet,itis,vos-wild} in which a user can provide both `positive' and `negative' clicks to indicate foreground and background/other objects respectively (as shown in Fig.~\ref{fig:outline}).  We denote the set of click positions as $\{\bm{p}_0, \bm{p}_1\}$ with subscripts 0 and 1 for positive and negative clicks respectively.  To date, guidance maps have been generated by as a function of the distance between each pixel of the image grid to the point of interaction.  More formally, for each pixel position $\bm{p}$ on the image grid, the pair of distance-based guidance maps for positive and negative clicks can be computed as \begin{equation}\label{eq:distg}
\mathcal{G}^d_{0}(\bm{p}) = \!\min_{\bm{c} \in \{\bm{p}_{0}\}}\!\! d(\bm{p}, \bm{c}) \;\; \text{and} \;\;
\mathcal{G}^d_{1}(\bm{p}) = \!\min_{\bm{c} \in \{\bm{p}_{1}\}}\!\!d(\bm{p}, \bm{c}).
\end{equation}
In the case of Euclidean guidance maps~\cite{ifcn}, the function $d(\cdot,\cdot)$ is simply the Euclidean distance; in Gaussian guidance maps, $d(\cdot,\cdot)$ is the value of a Gaussian with a standard deviation of $10$ pixels that is centred on the click~\cite{itis,vos-wild}.  Examples of such guidance maps can be found in Fig.~\ref{fig:motivation}.

However, such guidance is image-agnostic and assumes that each pixel in the scene is independent.  Our proposed approach eschews this assumption and proposes the generation of multiple guidance maps which align with both low-level and high-level image structures present in the scene.  We represent low-level structures with super-pixels and high-level ones with region-based object proposals and describe how we generate guidance maps from these structures in Sections~\ref{sec:superpixel} and ~\ref{sec:regionprop}.

\subsection{Superpixel-based guidance map}~\label{sec:superpixel} 
\noindent We first consider a form of guidance based on non-overlapping regions; in our implementation, we use superpixels. Superpixels group together locally similarly coloured pixels while respecting object boundaries~\cite{slic} and were the standard working unit of pre-CNN-based segmentation algorithms~\cite{fst,nlc}. Previous works have shown that most, if not all, pixels in a superpixel belong to the same category~\cite{nlc,fst,topdownbottomup}.  Based on this observation, we propagate user-provided clicks which are marked on single pixels to the entire super-pixel.  We then assign guidance values to each of the other super-pixels in the scene based on the minimum Euclidean distance from the centroid of each superpixel to the centroid of a user-selected super-pixel. One can think of the guidance as a discretized version of Eq.~\ref{eq:distg} based on low-level image structures.

\begin{figure*}[t!]
	\begin{center}
	\includegraphics[width=0.95\linewidth]{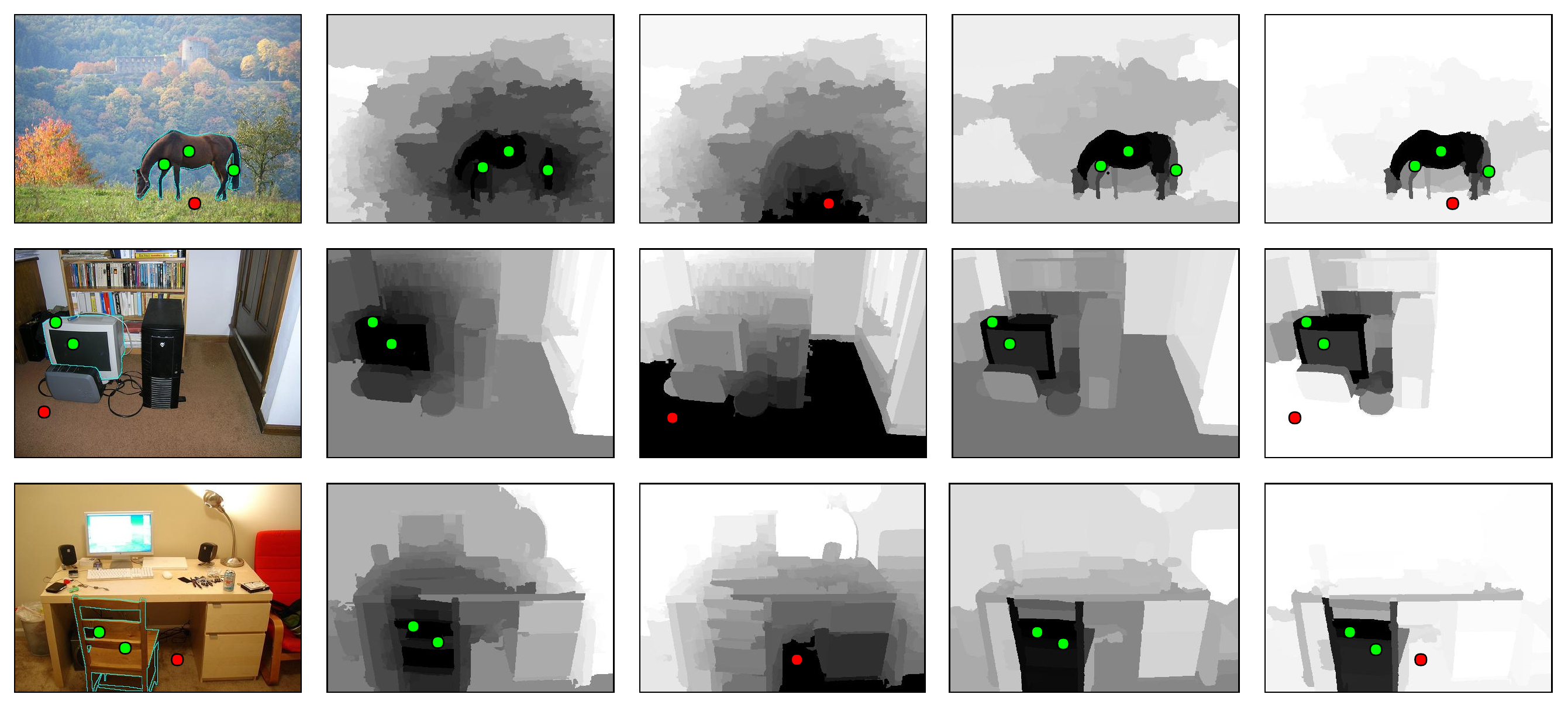}
	\end{center}
	\caption{\textbf{Example of guidance maps}. We transform the user-provided positive (shown as green dots) and negative (shown as red dots) clicks into guidance maps for the instance segmentation network (columns 2 to 5). The second and third column correspond to the positive and negative superpixel based guidance map respectively. Examples of the object based guidance map and the scale-aware guidance map are shown in columns 4 and 5 respectively. For the clarity of visualization, we inverted the values of the object based guidance map and the scale-aware guidance map (Best viewed in color).}
	\label{fig:samplemaps}
\end{figure*}

More formally, let $\{\mathcal{S}\}$ represent the set of superpixels partitioned from an image and $f_{SP}(\bm{p})$ be a function which maps each pixel location $\bm{p}$ in the image to the corresponding superpixel in $\{\mathcal{S}\}$. We further define a positive and negative superpixel set based on the positive and negative clicks, \ie $\{\bm{s}_0 = f_{SP}(\bm{p}_0)\}$ and $\{\bm{s}_1 = f_{SP}(\bm{p}_1)\}$ respectively. Similar to the distance-based guidance maps in Eq.~\ref{eq:distg}, we generate a pair of guidance maps.  However, rather than treating each pixel individually, we propagate the distances between superpixel centers to all pixels within each superpixel, \ie 

\begin{equation}
	\mathcal{G}^{\text{sp}}_t(\bm{p}) = \min_{s \in \{\bm{s}_t\}}\!\!d_c\left(s,f_{SP}(\bm{p})\right), \;\; \text{where}\;\; t=\{0,1\}, 
\end{equation}

\noindent and $d_c(s_i,s_j)$ is the Euclidean distance between the centers $s_i^c$ and $s_j^c$ of superpixels $s_i$ and $s_j$ respectively, where $s_i^c=(\sum_i x_i/|s_i|,\sum_i y_i/|s_i|)$ where $|s_i|$ denotes the number of pixels within $s_i$.  To maintain consistency across training images, the guidance maps values are then scaled to lie between $[0,255]$. In scenarios where the user provides no clicks, all pixel values are set to $255$.  Examples guidance maps are shown in the second and third column of Fig.~\ref{fig:samplemaps} respectively.

\subsection{Object-based guidance map}\label{sec:regionprop}

Super-pixels can be grouped together perceptually into category-independent object proposals. We also generate guidance maps from higher-level image structures, specifically region-based object proposals~\cite{ucm, geodesicproposal,cob,mcg,selectivesearch}. Such proposals have been used in the past as weak supervision for semantic segmentation~\cite{simple,boxsup} and allow us to incorporate a weak object-related prior to the guidance map, even if the instance is not explicitly specified by the user-provided clicks. To do so, we begin with a set of object proposals which have positive clicks its pixel support. For each pixel in the guidance map, we count the number of proposals from this set to which the pixel belongs. Pixels belonging to same object proposals are more likely to belong in the same object category and the number of proposals to which pixels belong incorporates a co-occurrence prior with respect to the current positive clicks.

More formally, let 
$\{\mathcal{L}_p\}$ be the set of object proposals for an image with support of pixel location $\bm{p}$. The object-based guidance map can be generated as follows: 
\begin{equation}\label{eqn:objectmap}
    \mathcal{G}^{\text{o}}(\bm{p}) = \sum_{\bm{p'} \in \{\bm{p}_0\}} \sum_{\mathcal{L} \in \{\mathcal{L}_{p'}\}} \mathbf{1}[\bm{p} \subset \mathcal{L}]
\end{equation}
where $\bm{1}[\bm{p}\!\subset\!\mathcal{L}]$ is an indicator function which returns 1 if object proposal $\bm{L}$ has in its support or contains pixel $\bm{p}$. Similar to the superpixel-base guidance map, the object-based guidance is also re-scaled  to $[0,255]$. In the absence of user-provided clicks, all pixels are set to the value of 0.  Examples  are shown in the third column of Fig.~\ref{fig:samplemaps}.

\subsection{Scale-aware guidance} 
Within an image, object instances can exhibit a large variation in their spatial extent~\cite{snip}. While deep CNNs are known for their ability to handle objects at different scales~\cite{deeplab}, specifying the scale explicitly leads to an improvement in performance~\cite{explicitscale}. Interactive instance segmentation methods~\cite{dextr} which isolate the object tend to have a superior performance. For segmenting object instances, it is thus desirable to construct guidance maps which exhibit spatial extents consistent with the object.

A common limitation of most click-based interactive approaches is that the provided guidance is non-informative about scaling of the intended object instance. The commonly used forms of guidance are either too localized~\cite{itis} (guidance map values are clipped to 0 at a distance of 20 pixels from the clicks) or non-informative~\cite{ifcn}.

Suppose now that we have some rough estimate of an object's scale in pixels, either in width or length.  A convenient way to make our guidance maps scale-aware is to incorporate contributions of superpixels and object proposals which are in agreement with this scale.  More specifically, we can apply this to the superpixel guidance map by truncating distances exceeding some factor $f$ of our scale measure $s$, \ie 
\begin{equation}
    \mathcal{G}^{\text{sp-sc}}_t(\bm{p}) = \max\left[\mathcal{G}^{\text{sp}}_t(\bm{p}), f s\right].
\end{equation}

We can apply similar constraints to the object-proposal based guidance by considering only the proposals within some accepted size range:
\begin{equation}
    \mathcal{G}^{\text{o-sc}}(\bm{p}) = \sum_{\bm{p'} \in \{\bm{p}_0\}} \sum_{\mathcal{L} \in \{\mathcal{L}_{p'}\}} \mathbf{1}[\bm{p} \subset \mathcal{L}] \cdot \mathbf{1}[f_1 \leq |\mathcal{L}| / s^2 \leq f_2].
\end{equation}

In the above equation, the second indicator function returns 1 only if the number of pixels in proposal $\mathcal{L}$ is bounded by some tolerance factors $f_1$ and $f_2$.

\subsection{Simulating user interactions}
Even when selecting the same object instance, it is unlikely that different users will provide the same interactions inputs.  For the model to fully capture expected behaviour across different users, one would need significant amounts of interaction training data.  Rather than obtaining these clicks from actual users for training, we simply simulate user clicks and generate guidance maps accordingly.

We follow the sampling strategies proposed in~\cite{ifcn}. For each object instance,  we sample $N_{pos}$ positive clicks within the object maintaining a distance $d^{in}_{1}$ pixels from the object boundary and $d^{in}_2$ pixels from each other. For negative clicks, we test the first two of the three sampling strategies outlined in~\cite{ifcn}, one in which $N_{neg}^1$ clicks are sampled randomly from the background, ensuring a distance of $d^{out}_1$ pixels away from the object boundary and $d^{out}_2$ pixels from each other and one in which $N_{neg}^2$ clicks on each of the negative objects (objects not of interest).

The above click-sampling strategy helps the network to understand notions such as negative objects and background but cannot train the network to identify and correct errors made during the prediction~\cite{itis}. To this end, we also randomly sample $N_{iter}$ clicks based on the segmentation errors.  After an initial prediction is obtained, positive or negative clicks are randomly sampled from the error. Existing set of clicks are then replaced with the newly sampled clicks with a probability of $0.3$. To mimic a typical user's behavior~\cite{itis}, the error-correction clicks are placed closest to the center of the largest misclassified region.

To estimate the scale measure $s$, we reserve the first two clicks, one positive and one negative, and assume that the Euclidean distance between the two is a roughly proportional measure; $f$, $f_1$ and $f_2$ are then set accordingly.


\section{Experimental Validation}

\subsection{Datasets \& Evaluation}
We apply our proposed guidance maps and evaluate the resulting instance segmentations on four publicly available datasets: PASCAL VOC 2012~\cite{pascal}, GrabCut~\cite{grabcut}, Berkeley~\cite{berkeley}, and MS COCO~\cite{mscoco}.

\noindent\textbf{PASCAL VOC 2012} consists of $1464$ training images and 1449 validation images spread across $20$ object classes.

\noindent \textbf{GrabCut} consists of $50$ images with the corresponding ground truth segmentation masks and is a used as a common benchmark for most interactive segmentation methods. Typically, the images have a very distinct foreground and background distribution.

\noindent\textbf{Berkeley} consists of $100$ images with a single foreground object. The images in this dataset represent the various challenges encountered in an interactive segmentation setting such as low contrast between the foreground and the background, highly textured background etc. 
	
\noindent\textbf{MS COCO} is a large-scale image segmentation dataset with $80$ different object categories, $20$ of which are from the Pascal VOC 2012 dataset. For fair comparison with~\cite{ifcn,risnet}, we randomly sample $10$ images per category for evaluation and splitting the evaluation for the $20$ Pascal categories versus the $60$ additional categories.
	
\noindent\textbf{Evaluation} Fully automated instance segmentation is usually evaluated with mean intersection over union (mIoU) between the ground truth and predicted segmentation mask.  Interactive instance segmentation is differently evaluated because a user can always add more positive and negative clicks to improve the segmentation and thereby increase the mIoU.  As such, the established way of evaluating an interactive system is according to the number of clicks required for each object instance to achieve a fixed mIoU.  Like~\cite{ifcn,risnet,itis,vos-wild}, we limit the maximum number of clicks per instance to 20. Note that unlike~\cite{ifcn,risnet}, we do not apply any post-processing with a conditional random field and directly use the segmentation output from the FCN.

\subsection{Implementation Details}

\paragraph{Training} As our base segmentation network, we adopt the FCN~\cite{fcn} pre-trained on PASCAL VOC 2012 dataset~\cite{pascal} as provided by MatConvNet~\cite{matconvnet}. The output layer is replaced with a two-class softmax layer to produce binary segmentations of the specified object instance. We fine-tune the network on the $1464$ training images with instance-level segmentation masks of PASCAL VOC $2012$ segmentation dataset~\cite{pascal} together with the $10582$ masks of SBD~\cite{sbd}. We further augment the training samples with random scaling and flipping operations. We use zero initialization for the extra channels of the first convolutional layer (\texttt{conv1\_1}). Following~\cite{ifcn}, we fine-tune first the stride-$32$ FCN variant and then the stride-$16$ and stride-$8$ variants. The network is trained to minimize the average binary cross-entropy loss.For optimization, we use a learning rate of $0.01$ and stochastic gradient descent with Nesterov momentum with the default value of $0.9$ is used.

\paragraph{Click Sampling} We generate training images with a variety of click numbers and locations; sometimes, clicks end up being sampled from the same superpixel, which reduces training data variation. To prevent this and also make the network more robust to the click number and location for training, we sample randomly from the following hyperparameters rather than fixing them to single values: $N_{pos}\!=\!\{2,3,4,5\}$, $N_{neg}^1\!=\!\{5,10\}$, $N_{neg}^2\!=\!\{3,5\}$, $d^{in}_{1}\!=\!\{15,20,40\}$, $d^{in}_2\!=\!\{7,10,20\}$, $d^{out}_1\!=\!\{15,40,60\}$, $d^{out}_2\!=\!\{10,15,25\}$. The randomness in the number of clicks and their relative distances prevents the network from over-fitting during training.

\paragraph{Guidance Dropout} Since the FCNs are pre-trained on PASCAL VOC $2012$, we expect the network to return a good initial prediction when it encounters image with object instances from one of its $20$ classes. Thus, during training, when the network receives images without any instance ambiguity (\ie an image with single object), we zero the guidance maps (value of $0$ for object guidance map and $255$ for the super-pixel based guidance map) with a probability of $0.2$ to encourage good segmentations without any guidance.  We further increase robustness by resetting either the positive or negative super-pixel based guidance with a probability of $0.4$.  
	
\paragraph{Interaction Loop}
During evaluation, a user provides positive and negative clicks sequentially to segment the object of interest. After each click is added, the guidance maps are recomputed; in addition the a distance transform of predicted mask from the previous iteration is provided as an extra channel~\cite{itis}. The newly generated guidance map is concatenated with the image and given as input to the FCN-$8$s network which produces an updated segmentation map.

\paragraph{Superpixels \& Object Proposals} We use the implementation provided in~\cite{mcg} for generating superpixels; on average, each frame has $500-1000$ superpixels. For comparison, we also try other superpixelling variants \eg SLIC~\cite{slic} and CTF~\cite{coarsetofine}.  Although several object proposal algorithms exist~\cite{selectivesearch,cpmc,mcg}, we use only MCG~\cite{mcg} as it has been shown to have higher quality proposals~\cite{boxsup}. The final stage of MCG returns a ranking which we disregard. We use the pre-computed object proposals for PASCAL VOC $2012$ and MS COCO provided by the authors of ~\cite{mcg}. For GrabCut and Berkeley, we run MCG~\cite{mcg} on the \emph{`accurate'} setting to obtain our set of object proposals. 

\subsection{Impact of Structure-Based Guidance} We begin by looking at the impact of super-pixel based guidance.  As a baseline, we compare with~\cite{ifcn}, which uses a standard Euclidean distance-based guidance as given in Eq.~\ref{eq:distg} (see examples in second row of Fig.~\ref{fig:motivation}). Similar to~\cite{ifcn}, we concatenate our positive and negative superpixel-based guidance maps with the three color channels and feed it as an input to the FCN-8s~\cite{fcn}. We use the superpixels computed using MCG~\cite{mcg}. For a fair comparison, we train our network non-iteratively, \ie, during training, we do not generate click samples based on the error in the prediction and do not append the distance transform of the current predicted mask as an extra channel. Looking at Table.~\ref{tab:grabcut}, we see that our super-pixel based guidance maps significantly reduce the number of clicks required to reach the standard mIoU threshold. 

The object-based guidance provides the network with a weak localization prior of the object of interest. adding the object-based guidance with the superpixel based guidance leads to further improvements in performance (see third row of Table.~\ref{tab:grabcut}). The impact is more prominent for datasets with a single distinct foreground object (\eg 9.3\% and 14\% relative improvement for the Berkeley and GrabCut dataset). Finally, by making the feedback iterative, \ie based on previous segmentation errors, we can further reduce the number of clicks.  Overall, our structure-based guidance maps can reduce the number of clicks by $35\%$ to $47\%$ and unequivocally proves that having structural information in the guidance map is highly beneficial. 

\begin{table}[ht!]
	\centering
	\begin{tabular}{cccc}
	\toprule
		& GrabCut  & Berkeley & VOC $2012$\\
			                     & @$90\%$  & @$90\%$  & @$85\%$\\
			                                \midrule
		Euclidean (\cite{ifcn}) & 6.04 & 8.65 & 6.88 \\                 
		SP          & 4.44     & 6.67 & 4.23\\
		    SP + Obj.  & 3.82     & 6.05 & 4.02\\
			\midrule
			SP + Obj. + Iter  & \textbf{3.58}     & \textbf{5.60} &\textbf{3.62}\\ 
	\bottomrule
	\end{tabular}
	\vspace{5pt}
	\caption{Clicks required for different types of guidance.  Guidance maps leveraging structural information require significantly less clicks than Euclidean distance-based guidance.  
	\textit{SP} refers to the super-pixel guidance maps and \textit{Obj} refers to the obect based guidance map and \textit{Iter} refers to iterative training.}
	\label{tab:grabcut}
\end{table}

\subsection{Impact of Scale-Aware Guidance}
Due to fixed-size receptive field, FCNs experience difficulty when segmenting small objects~\cite{learndeconv}. The benefits of our scale-aware guidance map which is most pronounced for segmenting small objects; for large objects, it does not seem to much effect.  To highlight the impact of our guidance on small object instances, we pick the subset of $621$ objects (from PASCAL VOC $2012$) which are smaller than $32\times32$; objects smaller than this size are harder to identify~\cite{snip}. 
	 
\begin{figure}
	\hspace*{-1cm}    
		\centering
		\begin{minipage}[t]{.5\textwidth}
		\vspace{0pt}
		\centering
		\parbox{10cm}{
		\begin{tikzpicture}[scale=0.8]
		\begin{axis}[ylabel=$\text{Avg. number of clicks}$,xlabel=$f_2$,
		legend pos=south east,xtick={1.1,1.2,1.5,2.0,3.0,6.0},xmin=1,xmax=6,ymin=4.1,ymax=4.5,xmode=log,log ticks with fixed point,x post scale=1.1,xticklabel style = {rotate=90,anchor=east},ylabel near ticks,xlabel near ticks]
		\addplot+[smooth] coordinates
		{(1.1,4.38) (1.2, 4.372) (1.5,4.37) (2.0,4.385) (3.0,4.39) (6.0,4.47)};
		\addplot+[black, smooth, mark options={draw=black,fill=black}] coordinates
		{(1.1,4.21) (1.2, 4.27) (1.5,4.25) (2.0,4.3) (3.0,4.33) (6.0,4.45)};
		\addplot[mark=none, red, dashed]  coordinates {(1.0,4.48) (6.0,4.48)};
		\addlegendentry{Estimated scale}
		\addlegendentry{Ground truth scale}
		\addlegendentry{Scale agnostic  ($f_2=+\infty$,$f_1=0$)}
		\end{axis}
		\end{tikzpicture}}
		\begin{center}
		    \small{(a) Scale-Aware Guidance}
		\end{center}
		\end{minipage}%
		\begin{minipage}[t]{.4\textwidth}
		\vspace{0pt}				
		\centering
		\parbox{12cm}{
		\begin{tikzpicture}[scale=0.8]
		\begin{axis}[ylabel=$\text{Avg. number of clicks}$,xlabel=\#superpixels (in thousands),
		legend pos=south east,xtick={0.500,1,2,5,10},xmin=0.500,xmax=10,ymin=4.1,ymax=7.0,xmode=log,log ticks with fixed point,x post scale=1.1,xticklabel style = {rotate=90,anchor=east}, ylabel near ticks,xlabel near ticks]
		\addplot+[smooth, forget plot] coordinates
		{(0.500,4.45) (1, 4.29) (2,4.47) (5,5.08) (10,5.68)};
		\addplot[mark=none, red, dashed]  coordinates {(0.500,6.88) (10,6.88)};
		\addlegendentry{iFCN}
		\end{axis}
		\end{tikzpicture}}
		\begin{center}
		    \small{(b) Number of superpixels}
		\end{center}
		\end{minipage}
		\caption{\textbf{(a) Scale-Aware Guidance.} The figure shows the average number of clicks required for segmenting small object instances (smaller than $32\times 32$ pixels~\cite{snip}) for varying degrees of tolerance till which we accept object proposals for generating our guidance map based on our estimated object scale and the ground truth object scale (computed as the square root of the number of pixels in the object mask). \textbf{(b) Number of superpixels.} The figure shows the average number of clicks required for segmenting object instances in PASCAL VOC 12 \textit{val} set for different number of superpixels.}
		\label{fig:combo}
	\end{figure}
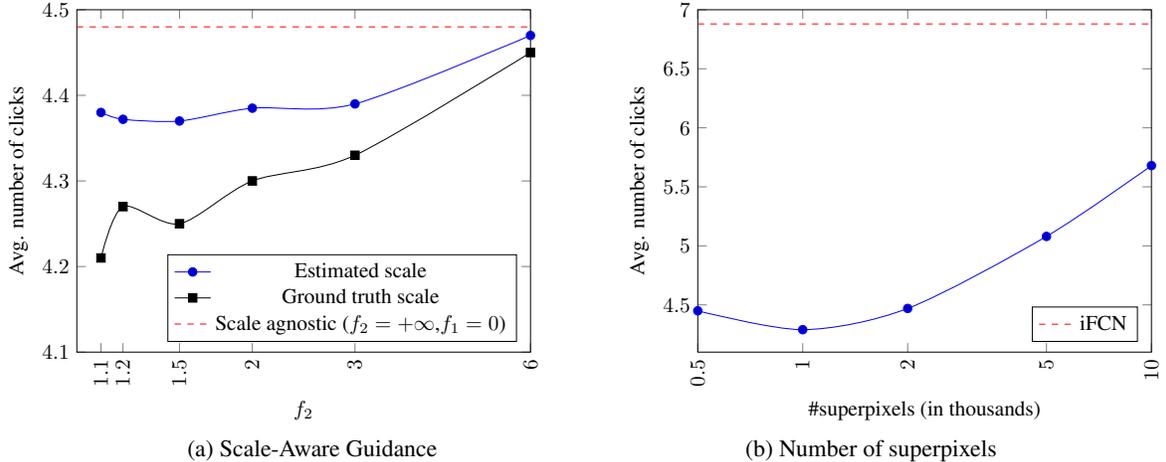

In the scale agnostic setting, we consider all object proposals which has the click in its pixel support for generating the object-based guidance map, \ie (as shown in  Equation.~\ref{eqn:objectmap}; note that this is equivalent to having $f_1\!=0, f_2\!=\!\infty$).  Since the lower bound on scale has little effect, we set $f_1\!=\!0$.  Looking at the average number of clicks required per instance to reach 85\% mIoU for the subset of small objects (see Fig.~\ref{fig:combo} (a)), we find that having a soft scale estimate improves the network performance when it comes to segmenting smaller objects.  This is primarily because the guidance map disregards object proposals which are not consistent in scale and can degrade the network performance by inducing a misleading co-occurrence prior.

When the scale $s$ is based on ground truth (as the square root of the number of pixels in the mask foreground, see blue curve in Fig.~\ref{fig:combo} (a)), the average clicks required per instance is consistently lower than the scale-agnostic case, even when as we relax $f_2$ up to 6, \ie allowing for object proposals which are $6$ times larger than the actual object scale.  Estimating the scale from the clicks is of course much less accurate than when it is take from the ground truth masks (compare black curve vs blue curve in Fig.~\ref{fig:combo} (a)). Nevertheless, even with such a coarse estimate, we find improvements in the number of clicks required as compared to the scale-agnostic scenario (compare red dashed line in Fig.~\ref{fig:combo} (a)). When the scale $s$ is based on the estimated scale (see blue curve in Fig.~\ref{fig:combo} (a)), the average clicks required per instance is consistently lower than the scale-agnostic case, even when as we relax $f_2$ up to 6, \ie allowing for object proposals which are $6$ times larger than the estimated object scale. Given the first pair of positive and negative clicks, our estimated object scale is $\sqrt{\pi}d$ where $d$ is the euclidean distance between the positive and the negative click. In our experiments, we observed that our estimated scale varies between $50-300\%$ from the ground truth scale). 
The difficulty of segmenting small objects using CNNs have been reported in literature~\cite{learndeconv}. Based on our preliminary experiments, we also observed that segmenting smaller objects can be problematic. For objects smaller than $32\times 32$ pixels from PASCAL VOC $2012$ \textit{val} set, we require an average of $4.33$ clicks which is significantly higher than our dataset average of $3.62$ clicks. At this point, we cannot  compare to existing state-of-the-art interactive segmentation approaches as these approaches typically report the average number of clicks over the entire dataset and do not distinguish between smaller and larger objects.

\subsection{Comparison to State of the Art}
\begin{figure}
	\centering
	\begin{minipage}{.5\textwidth}
	\centering
		\begin{tikzpicture}
		\begin{axis}[ylabel=$\text{mIoU}$,xlabel=$\text{Number of clicks}$,
		legend pos=south east,xmin=0,xmax=20,ymin=0.5,ymax=1,ylabel near ticks,xlabel near ticks]
		\addplot+[no marks,smooth] coordinates
			{(1,0.53573) (2,0.63382) (3,0.69688) (4,0.75223) (5,0.77911) (6,0.7974) (7,0.82285) (8,0.8351) (9,0.85559) (10,0.85738) (11,0.86204) (12,0.87519) (13,0.88335) (14,0.8921)    (15,0.8905) (16,0.89625) (17,0.90147) (18,0.89738) (19,0.90696) (20,0.91054)};
		\addplot+[no marks,smooth] coordinates 
			{(1,0.71) (2,0.79) (3,0.836) (4,0.868) (5,0.887) (6,0.903) (7,0.915) (8,0.924) (9,0.931) (10,0.936) (11,0.94) (12,0.945) (13,0.948) (14,0.95) (15,0.952) (16,0.954) (17,0.956) (18,0.958) (19,0.959) (20,0.96)};
		\addplot+[no marks,smooth] coordinates 
			{(1,0.74) (2,0.81) (3,0.846) (4,0.882) (5,0.901) (6,0.912) (7,0.92) (8,0.932) (9,0.936) (10,0.938) (11,0.942) (12,0.945) (13,0.948) (14,0.952) (15,0.953) (16,0.956) (17,0.958) (18,0.961) (19,0.96) (20,0.962)};
		\addlegendentry{iFCN~\cite{ifcn}}
		\addlegendentry{ITIS~\cite{itis}}
		\addlegendentry{Ours}
		\end{axis}
		\end{tikzpicture}\\
		\begin{center}
		    \small{(a) Pascal VOC $2012$ \textit{val} set}
		\end{center}
	\end{minipage}%
	\begin{minipage}{.5\textwidth}
	\centering
		\begin{tikzpicture}
		\begin{axis}[ylabel=$\text{mIoU}$,xlabel=$\text{Number of clicks}$,
			legend pos=south east,xmin=0,xmax=20,ymin=0.5,ymax=1,ylabel near ticks,xlabel near ticks]
		\addplot+[no marks,smooth] coordinates
			{(1,0.62889) (2,0.74285) (3,0.79842) (4,0.83959) (5,0.83684) (6,0.8625) (7,0.88998) (8,0.88087) (9,0.88309) (10,0.91481) (11,0.91363) (12,0.92075) (13,0.92631) (14,0.91706) (15,0.93541) (16,0.93936) (17,0.95019) (18,0.93832) (19,0.95318) (20,0.94784)};
		\addplot+[no marks,smooth] coordinates 
			{(1,0.821) (2,0.84) (3,0.863) (4,0.882) (5,0.894) (6,0.902) (7,0.906) (8,0.921) (9,0.926) (10,0.933) (11,0.935) (12,0.94) (13,0.942) (14,0.944) (15,0.946) (16,0.949) (17,0.95) (18,0.951) (19,0.953) (20,0.954)};
		\addplot+[no marks,smooth] coordinates 
			{(1,0.832) (2,0.87) (3,0.892) (4,0.906) (5,0.912) (6,0.918) (7,0.921) (8,0.932) (9,0.936) (10,0.938) (11,0.942) (12,0.945) (13,0.948) (14,0.952) (15,0.953) (16,0.956) (17,0.958) (18,0.958) (19,0.959) (20,0.96)};
		\addlegendentry{iFCN~\cite{ifcn}}
		\addlegendentry{ITIS~\cite{itis}}
		\addlegendentry{Ours}
		\end{axis}
		\end{tikzpicture}\\
		\begin{center}
		    \small{(b) GrabCut}
		\end{center}
	\end{minipage}
	\caption{\textbf{mIoU \textit{vs} number of clicks} on the (a) Pascal VOC $2012$ \textit{val} set~\cite{pascal} and (b) GrabCut dataset~\cite{grabcut}.}
	\label{fig:miouclicks}
\end{figure}
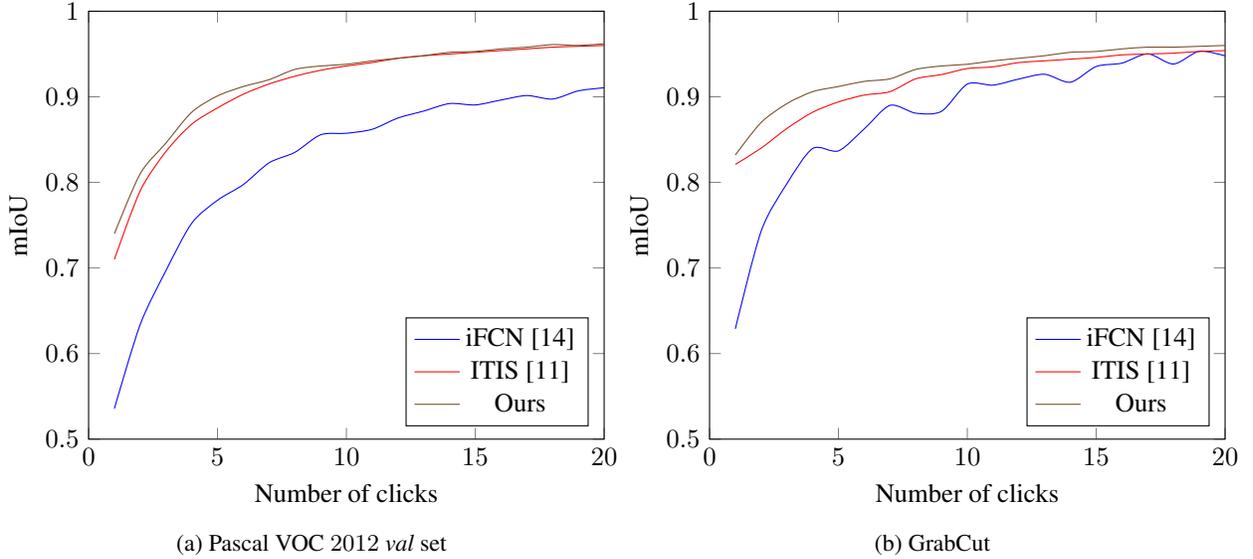

We compare the average number of clicks required reach some required mIoU (see Table~\ref{tab:all}) as well as mIoU \textit{vs.} the number of clicks (see Fig.~\ref{fig:miouclicks}) against other methods reported in the literature. The methods vary in the base segmentation network from the basic FCNs to the highly sophisticated DeepLabV3 and also make use of additional CRF post-processing.  We achieve the lowest number of clicks required for all datasets across the board, again proving the benefits of applying guidance maps based on existing image structures. We report results for our best trained SP+Obj+Iter network. In the initial stages of interaction on PASCAL VOC $2012$ \textit{val} set, our network outperforms the current state-of-the-art ITIS (as can be seen from Fig.~\ref{fig:miouclicks} (a)). 

To reach the mIoU threshold of $90\%$ on GrabCut and Berkeley, our full model needs the fewest number of clicks as shown in Table~\ref{tab:all} with a relative improvement of \textbf{5.79\%} and \textbf{7.13\%} over the current benchmark. For PASCAL VOC $2012$ \textit{val} set, we observe a relative improvement of \textbf{4.7\%}. For MS COCO, we observe a larger improvement for the $20$ seen categories from PASCAL VOC $2012$, as our networks were trained heavily on these object categories. Overall, we achieve an improvement of \textbf{9.7\%} and \textbf{5.28\%} over the $20$ seen and $60$ unseen object categories. We note that such an improvement is achieved despite the fact that our base network is the most primitive of the methods compared, \ie an FCN-8, in comparison to the others who use much deeper (ResNet-101) and more complex (DeepLab) network architectures. It should be noted that FCTSFN~\cite{twostream} and IIS-LD~\cite{latentdiversity} report their result over all the $80$ classes of MS COCO and not separately for $20$ seen classes and $60$ unseen classes.

\subsection{Superpixels}
\paragraph{Type of Superpixels} We consider two variants of superpixel, SLIC~\cite{slic} and CTF~\cite{coarsetofine} to study the impact of the type of superpixeling algorithm. For this study, we only consider the superpixel based guidance map and do not include the object based guidance map. On an average, MCG~\cite{mcg} generates $500-1000$ superpixels for each image in its default setting. For a fair comparison, we generate $500$ and $1000$ superpixels using SLIC and CTF. We observe that using 1000 SLIC superpixels results in performance similar to the MCG. However, irrespective of the superpixeling method, we found an overall improvement when the guidance maps are generated based on superpixels instead of pixel-based distances. 

\begin{table}[h]
	\centering
	\begin{tabular}{cccc}
	\toprule
		\#superpixels          & SLIC~\cite{slic} & CTF~\cite{coarsetofine} & MCG~\cite{mcg} \\ 
			\midrule
		500  & 4.45 & 4.82 & \multirow{2}{*}{4.23} \\
		1000 & 4.29 & 4.58 \\
	\bottomrule\\
	\end{tabular}
	\caption{Choice of superpixel algorithm}
	\label{tab:lowfeat}
	\vspace{-2pt}
\end{table}

\paragraph{Number of Superpixels} In this section, we study the impact of the number of superpixels. For this study, we only consider the superpixel based map as user guidance. We use SLIC~\cite{slic} as superpixel algorithm. In the extreme case, all superpixels will have one pixel in its support and the superpixel based guidance map degenerates to the Euclidean distance transform commonly used in existing interactive segmentation methods~\cite{ifcn,latentdiversity}. We use the reported results in iFCN~\cite{ifcn} on PASCAL VOC $2012$ \textit{val} set as our degenerate case (as shown by the red curve in Fig.~\ref{fig:combo} (b)). In addition to the reported results for $500$ and $1000$ superpixels on PASCAL VOC $2012$ \textit{val} set (as shown in Table 4 of the paper), we generate $2000,5000$ and $10000$ superpixels using SLIC~\cite{slic}. 

We notice an initial gain in performance, but with increase in the number of superpixels, the performance drops as our network requires more and more clicks to segment the object of interest. As the number of superpixels increase, the benefits of local structure based grouping is lost as each superpixel is segmented into similar and redundant superpixels.


\section{Qualitative Results}

\begin{figure*}[t!]
	\begin{center}
	\includegraphics[width=0.9\linewidth]{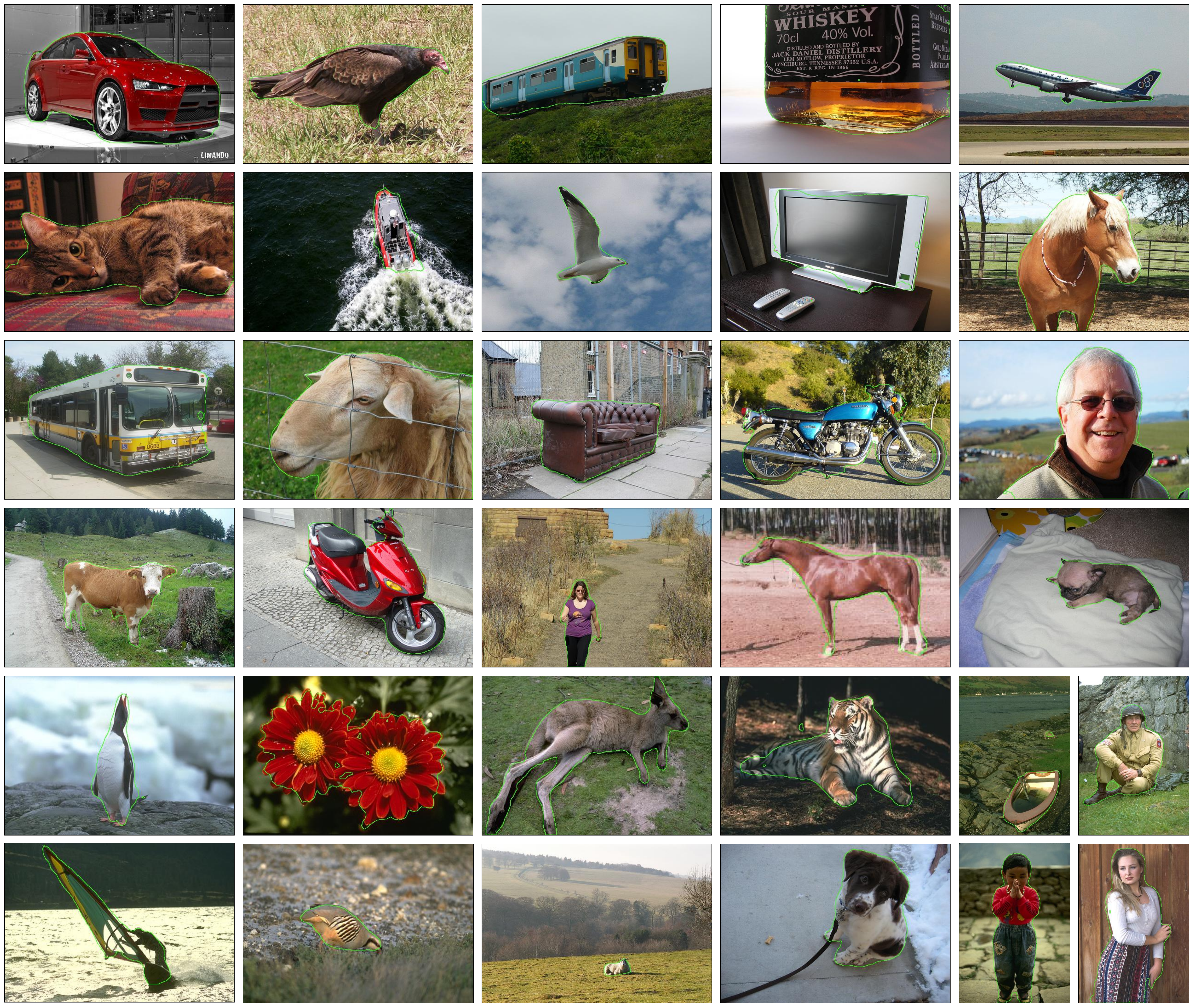}
	\end{center}
	\caption{\textbf{Zero Click Examples.} Examples of high-quality object masks generated without any user guidance.  In each example, there is only one foreground object and its appearance is very distinct from the background. Generated object boundaries are shown in green.  Figure best viewed in colour.} 
	\label{fig:zcberkeley}
\end{figure*}

\begin{figure*}[h]
	\begin{center}
	\includegraphics[width=0.9\linewidth]{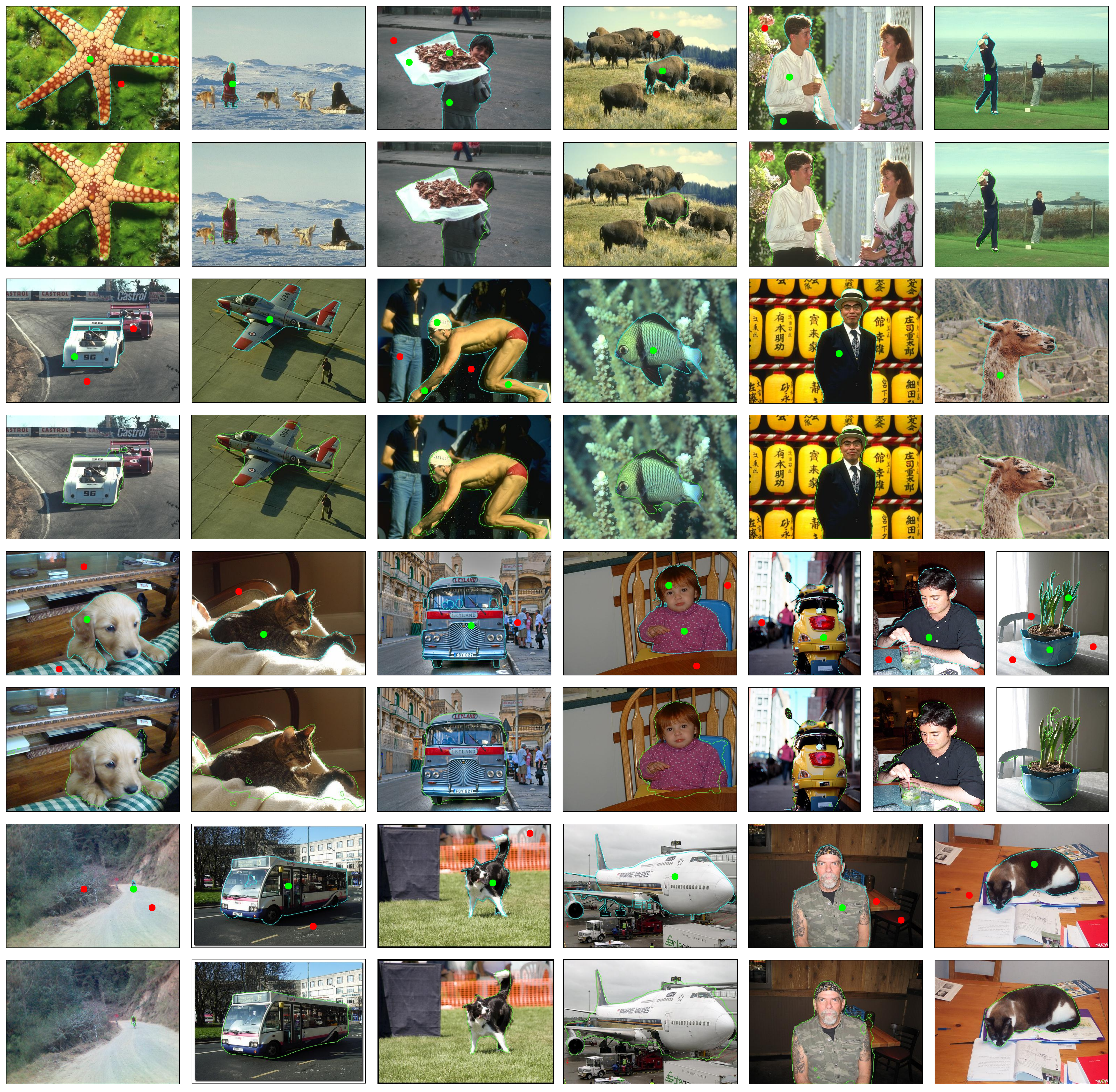}
	\end{center}
	\caption{\textbf{Multiple Click Examples.} With a few clicks, background and undesired object instances can be removed from the final prediction mask. Green dots indicate positive clicks and red dots indicate negative clicks. Ground truth object boundaries are shown in cyan and predicted object boundaries are shown in green.  Figure best viewed in colour.}
	\label{fig:multiclicks}
\end{figure*}

\begin{figure*}[t!]
	\begin{center}
	\includegraphics[width=0.9\linewidth]{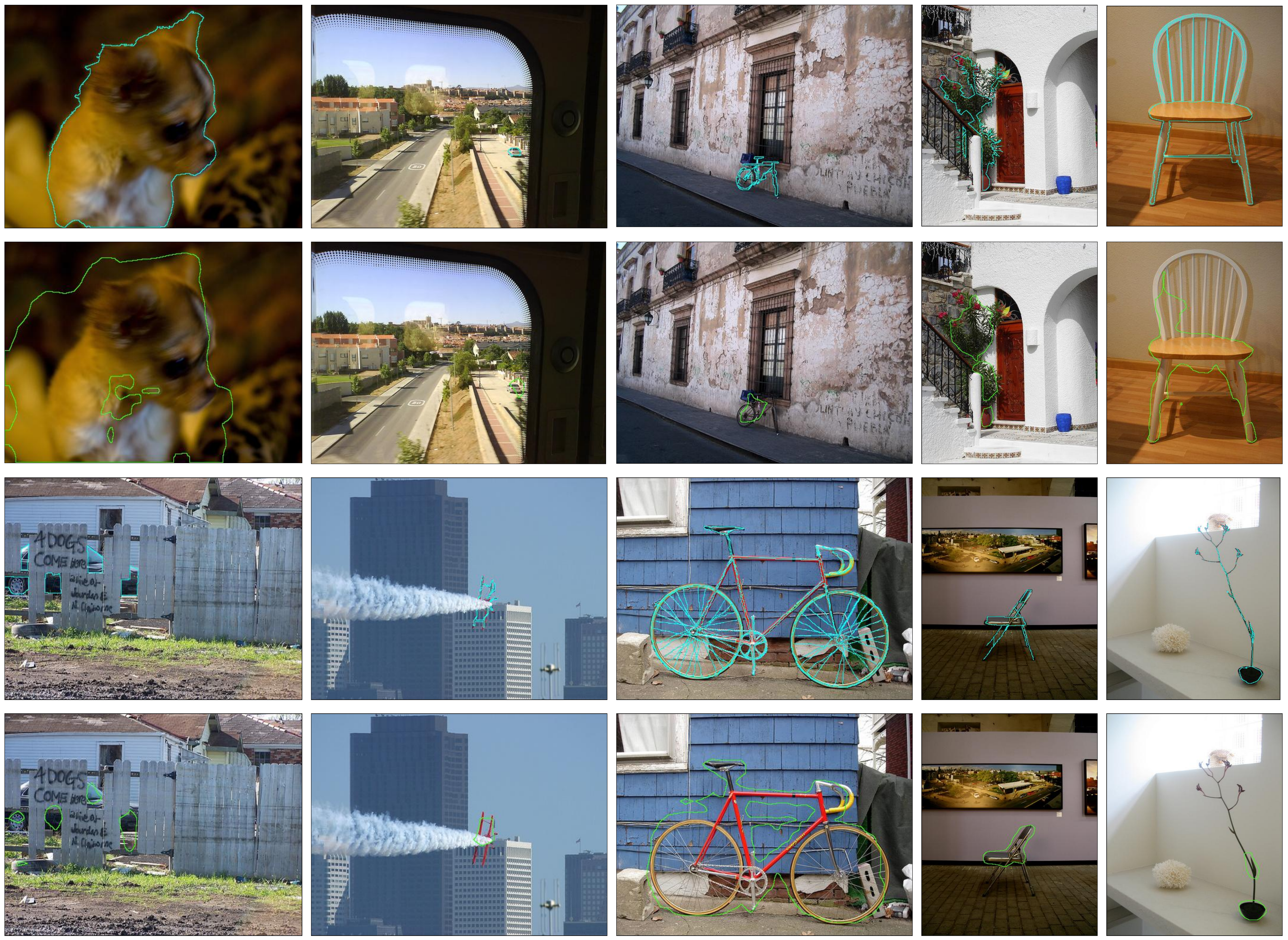}
	\end{center}
	\caption{\textbf{Failure Examples.} Ground truth object boundaries are shown in cyan while generated object boundaries from the predicted mask are shown in green. In the \textit{dog} example, the network has difficulty distinguishing the fur from the background. For the \textit{car} example, it is either too small (1st row, 2nd column) or too occluded (2nd row, 1st column). For the \textit{bicycle, chair} and \textit{potted plant} example, the error in prediction is due to the inability of the network in handling very fine structures. Figure best viewed in colour.} 
	\label{fig:20clicks}
\end{figure*}

\paragraph{Zero Click Examples}
We show via some qualitative examples, the benefits of having the guidance dropout. In several instances, our network is able to produce high quality masks without any user guidance (as shown in Fig.~\ref{fig:zcberkeley}).
\paragraph{Multiple Click Examples}
In Fig.~\ref{fig:multiclicks}, we show some examples where undesired objects and background was removed with only a few clicks resulting in a suitable object mask.
\paragraph{Failure Cases}
We show some examples from PASCAL VOC $2012$ \textit{val} set, where our network is unable to generate object masks with $\geq85\%$ mIoU and exhausts the $20$ click budget (see Fig.~\ref{fig:20clicks}). These failure cases are representative of the problems faced by CNNs while segmenting objects from images such as, small objects~\cite{learndeconv}, occlusion~\cite{amodal}, motion blur and objects with very fine structures. In general, we observed that our network had difficulty in handling three object classes from PASCAL VOC $2012$ - \emph{chair, bicycle} and \emph{potted plant}. This is primarily due to the inability of CNNs to produce very fine segmentations, most likely due to the loss of resolution from downsampling in the encoder.


\section{Discussion \& Conclusion}
In this work, we investigated the impact of the guidance maps for interactive object segmentation. Conventional methods use distance transform based approaches for generating guidance maps which disregard the inherent image structure. We proposed a scale aware guidance map generated using hierarchical image information which leads to significant reduction in the average number of clicks required to obtain a desirable object mask.

\begin{table}[t!]
	\centering
	\begin{tabular}{llccccc}
	\toprule
	Method         & Base   & GrabCut   & Berkeley & PascalVOC12   & MS-COCO     & MS-COCO\\
	& Network                & @$90\%$ & @$90\%$  & @$85\%$       & seen@$85\%$ & unseen@$85\%$ \\ 
	\midrule
	iFCN~\cite{ifcn}         & FCN-8s~\cite{fcn}             & 6.04 & 8.65 & 6.88 & 8.31 & 7.82\\
	RIS-Net~\cite{risnet}    & DeepLab~\cite{deeplab}             & 5.00 & 6.03 & 5.12 & 5.98 & 6.44\\
	ITIS~\cite{itis}         & DeepLabV3+~\cite{deeplabv3}   & 5.60 & - & 3.80 & - & - \\
	DEXTR~\cite{dextr}       & DeepLab~\cite{deeplab}    & 4.00 & - & 4.00 & - & - \\
	VOS-Wild~\cite{vos-wild} & ResNet-101~\cite{resnet}      & 3.80 & - & 5.60 & - & - \\
	FCTSFN~\cite{twostream}  & Custom                        & 3.76 & 6.49   & 4.58 & 9.62 & 9.62 \\
	IIS-LD~\cite{latentdiversity} & CAN~\cite{can} &  4.79 & -& -&12.45 &12.45 \\
	\textit{Ours}       & FCN-8s~\cite{fcn}             & \textbf{3.58} & \textbf{5.60} & \textbf{3.62} & \textbf{5.40} & \textbf{6.10}\\
	\bottomrule
	\end{tabular}
	\vspace{4pt}
	\caption{The average number of clicks required to achieve a particular mIoU score on different datasets by various algorithms. The best results are indicated in \textbf{bold}.}
	\label{tab:all}
\end{table}

During experimentation, we observed that the object instances within the datasets varied greatly in difficulty. For instance, on PASCAL VOC 2012, the base network, without \emph{any} user guidance, is able to meet the $\geq 85\%$ mIoU criteria for 433 of the 697 instances. Similarly observations were made for GrabCut ($\geq90\%$ mIoU, 13 out of 50) and Berkeley ($\geq90\%$ mIoU, 15 out of 100).  On the other hand, we encountered instances  where our algorithm repeatedly exhausted the $20$ click budget regardless of sampled click locations and iterative feedback based on prediction errors.  This is especially true for objects with very fine detailing, such as  such as spokes in bicycle wheels, partially occluded chairs, \etc. Based on these two extreme cases, we conclude that interactive segmentation is perhaps not so relevant for single object instances featuring prominently at the center of the scene and should feature more challenging scenarios.  On the other hand, we need to design better algorithms which can handle objects that are not contiguous in region, \ie has holes and are able to handle scenarios of occlusion. Depending on the target application, dedicated base architectures may be necessary to efficiently handle these cases. 


{\small
	\bibliographystyle{unsrt}
	\bibliography{ms}
}
\end{document}